%% file: bayesbias.tex
\documentclass{article}
\usepackage{amsmath}
\usepackage{amsthm}
\usepackage{amssymb}
\usepackage{mathtools}
\usepackage{epsf}
\usepackage{colt96}
\input{mathdefs}

\newenvironment{pf}{\noindent{\bf Proof.\ }}{$\Box$ \\ }

\begin{document}
\title{A Bayesian/Information Theoretic Model of Bias Learning}

\author{Jonathan Baxter \\
Department of Mathematics \\ London School of
Economics\\ and \\ Department of Computer Science \\ Royal Holloway College \\
University of London} 
\maketitle
{\sloppy
\begin{abstract}
In this paper the problem of learning appropriate bias for 
an environment of related tasks is examined from a Bayesian perspective. 
The environment of related tasks is shown to be naturally modelled by the 
concept of an {\em objective} prior distribution. Sampling from the objective
prior corresponds to sampling different learning tasks from the environment. 
It is argued that for many common machine learning problems, although we 
don't know the true (objective) 
prior for the problem, we do have some idea of a set of 
possible priors to which the true prior belongs. It is shown that under these 
circumstances a learner can use Bayesian inference to learn the true prior
by sampling from the objective prior.
Bounds are given on the amount of information required to learn a task when
it is simultaneously learnt with several other tasks. The bounds show that if the 
learner has little knowledge of the true prior, and the dimensionality of the 
true prior is small, then sampling multiple tasks is highly advantageous.
\end{abstract}
\section{Introduction}
In the VC and PAC models of learning \cite{VC1,VC2,Valiant}, and
indeed in most practical learning scenarios, the learner's bias is
represented by the choice of hypothesis space. This choice is
extremely important: if the space is too large the learner will not be
able to generalise well; if the space is too small it is unlikely to
contain a solution to the problem being learnt.

A desirable goal in machine learning is to find ways of 
automatically learning appropriate bias, rather than having to 
build the bias in by hand. In the VC context this means finding ways 
of automatically learning the hypothesis space. 
A VC-type model of bias learning in the context of learning internal representations
was introduced in \cite{colt95}, while a more general model that allows for 
any kind of specification of the hypothesis space is given in
\cite{BiasLearn95}. The central assumption of the model
is that the learner is embedded within an {\em environment} of related tasks. 
The learner is able to sample from the environment and hence generate multiple
data sets corresponding to different tasks. The learner can then search for
a hypothesis space that is appropriate for learning all the tasks. 
This model can be thought of as a first order approximation to the idea 
that when choosing 
an appropriate hypothesis space or model for a learning problem, we are 
doing so on the basis of experience of {\em similar problems}. 

It is
shown in \cite{colt95,BiasLearn95} 
that under certain mild restrictions on the set of all hypotheisis
spaces available to the learner, it is possible for the learner to 
sample sufficiently often from sufficiently many tasks to ensure that 
a hypothesis space containing hypotheses with small empirical loss on all the tasks will 
with high probability contain good solutions to novel tasks drawn from the 
same environment. It is also 
shown in those papers that 
if the learner is learning a common {\em internal representation}
or {\em preprocessing} for an $n$ task training set (see figure 
\ref{nnet}) then the number
of examples $m$ required of each task to ensure good generalisation obeys
\begin{equation}
\label{eq1}
m = O\(a + \frac{b}{n}\).
\end{equation}
Here $a$ is a measure of the dimension of the smallest hypothesis space needed
to learn all the tasks in the environment and $b$ is a measure of the 
dimension of the space of possible preprocessings available to the learner.
The $n=1$ case of formula \eqref{eq1} 
is an upper bound the number of examples 
that would be required for good generalisation in the ordinary, single 
task learning scenario, while the limiting case of $m=O(a)$ is an upper 
bound on the number of
examples required if the correct preprocessing is already known. Thus, this
formula shows that the upper bound on the number of examples required 
per task for good generalisation decays to the minimal possible as the number 
of tasks being learnt increases. 

Although very suggestive, without a matching
lower bound of the same form, we cannot actually conclude from 
\eqref{eq1} that learning multiple related tasks 
requires fewer examples per task for good generalisation than if those 
tasks are learnt independently. 
Unfortunately, lower bounds within a real-valued VC/PAC framework
are in general very difficult to come by because  
an infinite amout of information can be conveyed
in a single real value and so it is possible to construct complicated 
function classes in which the identity of each function is encoded in its 
value at every point (see \eg \cite{BLW}). This suggests that rather
calculating the number of {\em examples} required to learn, we should 
calculate the amount of {\em information} required to learn. 

In this paper the  model of bias learning introduced in 
\cite{colt95} and \cite{BiasLearn95} is modified to 
a Bayesian model of bias learning. There are a number of reasons for this.
One is that the question ``how much information is required to learn'' 
is more natural within a Bayesian model than within the VC model. 
Another reason is that it is much easier to formulate and analyse the
effects of prior knowledge on the learning process. This is particularly 
important in bias learning where we are trying to understand how the process of
aquiring prior knowledge can be automated. In the VC framework the learner's 
prior knowledge is represented by the hypothesis space chosen for 
the problem. All hypotheses within the hypothesis space are viewed equally,
whereas in the Bayesian framework the learner can rank the hypotheses in
order of prior preference using a prior distribution. In addition, the Bayesian learner 
does not have to choose a particular hypothesis as the result of the learning process,
it simply ranks the alternative hypotheses in the light of the data. Finally, quantities
involving {\em information} (in the Shannon sense) have a more natural expression 
within a Bayesian framework. 

The main feature of the Bayesian {\em bias learning} model 
introduced here is that the 
prior is treated  as {\em objective}. The sample space of the prior 
represents the space of tasks in the environment, and sampling from the
prior corresponds to selecting different learning tasks 
from the environment. The analagous question
to ``how many examples are required of each task in an $n$ task training 
set'' leading to the upper bound \eqref{eq1}, is ``how much information
is required per task to learn $n$ tasks?'' We will see that if the learner 
already knows the true prior then there is no advantage to learning $n$ tasks;
that is, the expected amount of information needed to learn each task within
an $n$ task training set is the same as if the tasks are learnt separately.
However, if the learner does not know the true prior (which is
generally the case in bias learning, otherwise there is no need to do 
bias learning), but instead knows only
that the prior is one of a set $\Pi$ of possible priors (the possible priors
in this case correspond to the different hypothesis spaces available to the 
learner in the VC/PAC model of bias learning), 
then we will see that the expected information
needed per task, $\Rbar_{n,\pi^*}$, obeys asymptotically (in $n$)
\begin{equation}
\label{eq2}
\Rbar_{n,\pi^*}  \thickapprox a' + b'(\pi^*)\frac{\log n}n + o\(\frac{\log
n}{n}\)
\end{equation}
where $a'$ is the minimal amount of information possible (the amount the learner would require
if it knew the true prior $\pi^*$) and $b'(\pi^*)$ is a local
measure of the dimension of the 
space of possible priors $\Pi$ at the point $\pi^*$. 
Here $f(n,\pi^*) \thickapprox g(n,\pi^*)$  means $f(n,\pi) = g(n,\pi)$ for all 
but a set of $\pi$ of vanishingly small measure as $n\rightarrow\infty$.
Comparing \eqref{eq2} and \eqref{eq1} and the 
meaning of $a$ and $b$ with their partners $a'$ and $b'$, we see
that this partially realises the aim of providing an exact bound justifying
learning multiple related tasks.

The question of how much information is required to encode the $m$'th
observation of each task in an $n$ task training set is also analysed in this
paper, and an example is given showing that when the true prior is unknown,
learning multiple tasks is also highly advantageous in this setting.

The rest of the paper is organised as follows. The Bayesian model
of bias learning is introduced formally in section \ref{bbsec}, along
with a concrete example based on neural networks for image recognition.
The relationship between Bayesian bias learning as formulated here and 
{\em hierarchical} Bayesian methods is also discussed.
Equation \eqref{eq2} is derived in section \ref{results} and the constants
$a$ and $b$ are calculated for the neural network example, where again
contact is made between the Bayesian model results and the VC model results.
In section \ref{nmsec} the question of how much information is required 
to encode the $m$'th observation of each task in an $n$ task training set is 
analysed. In section \ref{dimsec} the dimension of a fairly general class
of smoothly parameterised models is calculated, leading to a
characterisation  of the advantages of multiple task learning within a
Bayesian context.

\subsection{Notation}
The probability model treated throughtout this paper is three-tiered. At the
bottom level is $Z$ which is assumed to be a complete separable metric space.
All probability measures on $Z$ are defined on the sigma-field of Borel
subsets of $Z$. $Z$ is the learner's interface with the environment---the
learner receives all its data in the form of samples from $Z$. The next
level up in the hierarchy is $\Theta$, which is the set of possible
``states of nature'' or ``learning tasks'' with which  
the learner might be confronted. For each $\theta\in\Theta$ there is
a probability measure $P_{Z|\theta}$ on $Z$. We assume there exists a fixed
$\sigma$-finite measure $\nu$ that dominates $P_{Z|\theta}$ for each 
$\theta\in\Theta$. $\Theta$ is also assumed to be a complete separable metric
space. At the highest level in the hierarchy is the set 
$\Pi$ which 
represents the space of possible ``priors'' on $\Theta$. For each $\pi\in\Pi$ 
there is a probability measure $P_{\Theta|\pi}$ on $\Theta$. Again the
$P_{\Theta|\pi}$'s are defined on the sigma field of Borel subsets of
$\Theta$ and we assume there exists a second measure $\mu$ dominating all 
$P_{\Theta|\pi}$. 
Finally, on $\Pi$ 
there is a fixed probability measure $P_\Pi$: the ``hyper-prior''. As
$\Theta$ is a complete separable metric space, we can take the domain of
$P_\Pi$ to be the sigma field generated by the topology of weak convergence 
of the $P_{\Theta|\pi}$ measures. 

Integration with respect to the measures $\nu$ and $\mu$ will be denoted by
$\int_Z \, dz$ and $\int_\Theta\, d\theta$ respectively ($\nu$ and $\mu$ are not
assumed to be Lebesgue measures---the notation is just for convenience).
Integration with respect to the hyper-prior $P_\Pi$ will be denoted
$\int_\Pi p(\pi)\, d\pi$
The Radon-Nikodym derivative of any measure $P_{Z|\theta}$ at $z\in Z$,
$\frac{dP_{Z|\theta}}{d\nu}(z)$ will be written interchangably as $p(z|\theta)$ or
$p_{Z|\theta}(z)$, and similarly $\frac{dP_{\Theta|\pi}}{d\mu}(\theta)$ will be
written as $p(\theta|\pi)$ or $p_{\Theta|\pi}(\theta)$.

If $f$ is a function on $Z$, then the expectation of $f$ with respect to any 
random variable with 
distribution $P_{Z|\theta}$ will be denoted by 
$\E_{Z|\theta} f(z) = \int_Z f(z) p(z|\theta)\, dz$. Similarly for functions
defined on $\Theta$ and $\Pi$. 

$n\times m$ matrices with elements from $Z$ will be denoted by $\z$:
$$
\z =
\begin{matrix}
z_{11} & \hdots & z_{1m} \\
\vdots & \ddots & \vdots \\
z_{n1} & \hdots & z_{nm}.
\end{matrix}
$$
The columns of $\z$ will be denoted as $z^n_i$, so $\z = [z^n_1\dots z^n_m]$.

Let $\N$ denote the natural numbers.

\section{The Basic Model}
\label{bbsec}
In Bayesian models of learning (see \eg \cite{Berger86}) the learner recieves data
$z^n = z_1,\dots,z_n$ which are observations on $n$ random
variables $Z^n = Z_1,\dots,Z_n$. The $Z_i$ are identically distributed and 
conditionally independent 
given the true state of nature $\theta$. The learner does not know $\theta$,
but does know that $\theta$ belongs to a set of possible states of nature 
$\Theta$. The learner begins with a prior distribution $p(\theta)$
and upon receipt of the data 
$z^n$ updates $p(\theta)$ to a posterior distribution $p(\theta|z^n)$
according to Bayes' rule:
\begin{equation}
\label{basicbrule}
p(\theta|z^n) = \frac{p(z^n|\theta) p(\theta)}{p(z^n)},
\end{equation}
where
$$
p(z^n) = \int_\Theta p(z^n|\theta) p(\theta)\, d\theta.
$$

Bayesian approaches to neural network learning have been around for a while
(see \eg \cite{Mackay}), and they essentially constitute a subset 
of Bayesian approaches to non-linear regression and classification.  
Mapping these approaches on to the present framework, consider 
the case of an MLP for recognising my face.
The weights of the network correspond to 
the set of possible states of nature $\Theta$, the true state of nature 
$\theta^*$ being an assignment of weights such that the output of the 
network is 1 when an example of my face is applied to its input,
and 0 if anything else is applied to its input. The data $z^n = z_1,\dots,z_n$
comes in the form of input-output pairs $z_i = (x_i,y_i)$ where each $x_i$ 
is an example image and $y_i$ is the correct class label (in this case either
0 or 1). Note that as we are only interested in classification in this 
example, the input distribution
$p(x)$ is not modelled, only the conditional distribution 
on class labels $p(y|x)$. Denoting the output of the network by $f_\theta(x)$,
and interpreting $f_\theta(x)$ as $p(y=1|x)$,
it can easily be shown \cite{Bridle} that the
probability of data set $z^n = (x_1,y_1),\dots,(x_n,y_n)$ given weights  
$\theta$ is 
\begin{equation}
\label{zog}
p(z^n|\theta) = \prod_{i=1}^n p(x_i) e^{-E(z^n;\theta)}
\end{equation}
where
$$
E(z^n;\theta) = \sum_{i=1}^n y_i \log(f_\theta(x_i)) + 
(1-y_i)\log(f_\theta(x_i)).
$$
Choosing a prior (typically multivariate Gaussian or uniform over some 
compact set) for the weights and 
substituting \eqref{zog} into \eqref{basicbrule} yields the posterior 
distribution on the weights $p(\theta|z^n)$. The posterior is the 
``output'' of the learning process. It can be used to predict the class 
label of a novel input $x^*$ by integrating:
$$
p(y=1|x^*;z^n) = \int_\Theta f_\theta(x^*) p(\theta|z^n)\, d\theta.
$$
\subsection{Interpreting the Prior} 
In the example above the prior $p(\theta)$ is a purely 
{\em subjective} prior. As is typical for these problems a
relatively weak prior is chosen reflecting our weak 
knowledge about appropriate weight settings for this problem.
However, in the case of face recognition (and many other pattern recognition
problems such as speech and character recognition) 
it is arguable that there exists 
an {\em objective} prior for the problem. To see this, note that given our
weak prior knowledge we are likely to have chosen a network large enough
to solve {\em any} face recogition problem within some margin of 
error, not just the specific task:
``recognise Jon ''. Hence it is likely that 
there will exist weight settings
$\theta_1, \theta_2, \theta_3, \dots$ that will cause the network to 
behave as a classifier for
`Mary', `Joe', `males', `smiling', `big nose' 
and so on. In fact there should exist weight
settings that correspond to nonexistent faces 
provided different examples of the face vary in a ``face-like'' way.
Hence we can consider the space of all face classifiers, 
both real and fictitious, as represented by a particular subset 
$\Theta_{\text{face}}$ of all possible weight settings 
$\Theta$. The {\em objective prior} $p(\theta)$ for face recognition is then
characterised by the fact that its support is restricted to  
$\Theta_{\text{face}}$. The restriction of the support is the most
important aspect of the face prior. The actual numerical probabilities for each
element $\theta\in\Theta_{\text{face}}$ could be chosen in a number 
of different ways, but for the sake of argument we can take them to be 
uniform or as corresponding to the general frequency of face-like classifier 
problems encountered in a particular person's environment.

The usual subjective priors chosen in neural network applications 
(Gaussian or uniform 
on the weights) bear no resemblence to the objective prior discussed above:
initializing the weights of a network according to a Gaussian prior typically
does not cause the network to behave like some kind of face classifier, 
whereas initializing according to the objective prior by definition will
induce such behaviour. Hence the use of subjective priors such as the Gaussian  
not only demonstrates our ignorance concerning the specific task at hand
(\eg learn to recognise Jon) but also demonstrates our ignorance concerning the
true prior. That is, we typically have little idea which parameter settings 
$\theta$ correspond to face-like classifiers and which correspond to 
``random junk''. 

Should we care that we don't know the true prior? In short: yes. If we 
know the true prior then the task of learning any individual face is
vastly simplified. A single positive example of my face is enough to set
the posterior probability  of any other individual face classifiers to 
zero (or very close to zero),
and a few more examples with me smiling, frowning, bearded, clean-shaven,
long-haired, short-haired and so on is enough to set the posterior probability
of {\em every} other classifier (the smiling, frowning, \etc classifiers) 
except the ``Jon'' classifier to zero. Contrast this with the usual subjective
priors where
typically thousands of examples and counter-examples of my face would have to
be supplied to the network before a reasonably peaked posterior and hence
reasonable generalisation could be achieved. 

\subsection{Learning the Prior}
\label{learnprior}
If knowing the true prior is such a great advantage then we should try to 
learn it. To do this we can set up a space of candidate priors indexed 
by some set $\Pi$. Thus, each $\pi \in \Pi$ corresponds to some  
prior $p(\theta|\pi)$ on $\Theta$. We assume realizability, so that 
the objective prior $p(\theta|\pi^*)$ corresponds to some $\pi^*\in\Pi$.
To complete the Bayesian picture a {\em subjective} prior 
$p(\pi)$ must be chosen for $\Pi$. Typically we will not have a strong 
preference for any particular prior 
and so we can follow the course taken in ordinary Bayesian inference
under such circumstances and choose $p(\pi)$ to be non-informative or 
simply Gaussian with large variance or uniform over some compact set
(assuming $\Pi$ is Euclidean).
 
As the true prior $p(\theta|\pi^*)$ is objective we can 
{\em in principle} sample from 
it 
to generate a sequence of training {\em tasks}\footnote{In reality 
we cannot sample directly from the 
prior to get $\theta_1,\theta_2,\dots$, only from conditional 
distributions $p(z|\theta_1), p(z|\theta_2), \dots$. This is 
discussed further in section \ref{nmsec}. For the moment we maintain the 
fiction that we have direct access to the parameters $\theta$.} $\theta^n = \theta_1,\theta_2,\dots,
\theta_n$. A direct application of Bayes' rule then gives the posterior
probability of each prior:
$$
p(\pi|\theta^n) = \frac{p(\theta^n | \pi) p(\pi)}{p(\theta^n)}
$$
where $p(\theta^n|\pi) = \prod_{i=1}^n p(\theta_i|\pi)$ and 
$p(\theta^n) = \int_\Pi p(\theta^n|\pi) p(\pi)\, d\pi$.

Under appropriate conditions the posterior 
distribution will tend to a delta function over the true prior $\pi^*$ as 
$n\rightarrow \infty$. Thus for large enough $n$ the learner can be said to 
have {\em learnt the prior}. 

For this model to work we have to assume that although the 
learner has no idea about the true prior, it can generate a class of priors
$\Pi$ containing the true prior $\pi^*$. This assumption is quite reasonable 
in the case of face recognition because it seems plausible that there 
exists a {\em low-dimensional internal representation} for faces such 
that each face classifier can be implemented by a simple map (\eg linear 
or nearest-neightbour) composed with the internal representation. A low 
dimensional representation (LDR) in its simplest form is just a fixed mapping
from the (typically high-dimensional) input space to a much smaller 
dimensional space. One can think of the LDR as a preprocessing applied to the
input data that extracts features that are important for classification.
For example, in the case of
face recognition it might be that to uniquely determine any face one only
needs to know the distance between the eyes and the length of the nose.
So an appropriate LDR would be a two-dimensional one that extracts these 
two features from an image. Although faces almost certainly cannot
be represented solely 
by the inter-eye distance and nose length,
it is highly plausible that 
some kind of LDR exists for the face recognition problem. It is similarly 
plausible that LDR's exist for other pattern recognition problems such as 
character and speech recognition\footnote{The ability of humans to learn to
recognize spoken words, written characters and faces with just a handful of 
examples indicates that some kind of LDR must be employed in our processing.
Even if our internal representations are not strictly lower dimensional than
the raw input representation, the maps we compose with our internal
representations must be very ``simple'' in order for us to learn with so 
few examples.}

Figure \ref{nnet} illustrates how in the case of neural-network
learning the assumption that there exists an LDR for the tasks in the
environment can be translated into a specification for the set of
possible priors $\Pi$. The hidden layers of the network labelled LDR correspond to
the LDR, while each individual classifier task is assumed to be
implementable by composing a linear map with the output of the LDR.
Thus each $\theta\in\Theta$ divides into two parts: $\theta =
(\theta_{\text{LDR}} , \theta_{\text{OUT}})$, where $\theta_{\text{LDR}}$
are the hidden layer weights and $\theta_{\text{OUT}}$ are the weights
of the linear output map.  As a first
approximation, it is reasonable to assume that the true prior
$p(\theta|\pi^*)$ is a delta function positioned at 
$\theta^*_{\text{LDR}}$---the true preprocessing (LDR), and fairly
uniform over output layer weight settings. Hence it is reasonable to
take $\Pi$ to be the set of all priors that are a delta function
over some $\theta_{\text{LDR}}$, and fairly smooth Gaussians
(or uniform distributions) over $\theta_{\text{OUT}}$. To simplify
matters assume that the 
distribution on $\theta_{\text{OUT}}$ is the same for all priors. With
these assumptions, $\Pi$,  the set of possible priors of this form
is isomorphic to the set of possible weights in the hidden
layers, $\Theta_{\text{LDR}}$. In this model knowing the true prior is
equivalent to knowing the correct input-hidden layer weights. Learning any
individual task is then simply a matter of estimating the output weights for
a single node which is a simple problem of linear regression. The output layer
weights are thus the model parameters while the hidden layer weights are the 
model hyper-parameters.
\begin{figure}
\begin{center}
\leavevmode
\epsfysize=2.5in\epsfbox{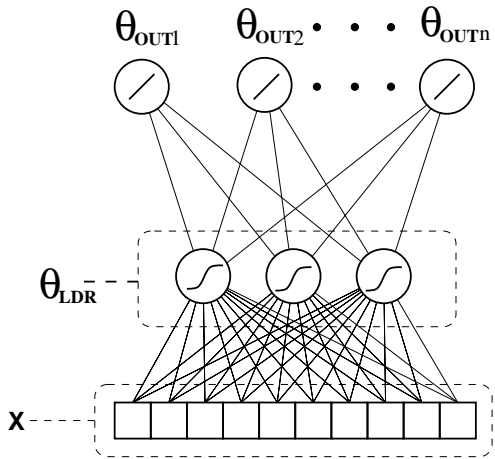}
\caption{\label{nnet}A neural network for low dimensional representation 
(LDR) learning.
Each task in the environment is implemented by composing a linear map with 
weights 
$\theta_{\OUT i}$ with a fixed preprocessing or LDR. In the example
considered in this paper the LDR is a single layer neural network with
sigmoidal nodes. The weights of the LDR are $\theta_\LDR$. The $\theta_\LDR$
weights are {\em hyper-parameters} while the $\theta_\OUT$ weights are 
ordinary model parameters.}
\end{center}
\end{figure}

\subsection{Relationship to hierarchical Bayes and existing Bayesian 
neural network techniques}
The framework outlined in the previous section is in fact a special case
of what is known as {\em hierarchical Bayesian inference} 
(see e.g \cite{Berger85,Berger86,Good80}). Hierarchical Bayesian inference has also 
been discussed in the context of neural networks by several authors 
(see \eg \cite{Mackay}, although the techniques presented there are not
explicitly identified by the author as hierarchical Bayes). 
The distinction 
between {\em subjective} and {\em objective} priors has been observed and 
the idea of multiple sampling from objective priors has been analysed for
a number of different models. However the models analysed are typically 
quite low-dimensional in comparison to the kind of models used in 
neural network research. Will we see in the following section that 
Hierarchical Bayes with multiple task sampling can be particularly useful
in high-dimensional models.

To the best of my knowledge the idea of an objective prior has not been employed
previously in Bayesian approaches to neural networks. For the most part 
the hierarchical Bayes approach has been used to 
tune a small number of ``nuisance'' (hyper) parameters (such as the parameter 
$\lambda$ controlling the trade-off between regularisation and data-misfit 
in regression networks \cite{Mackay2}). Note that these are the only 
parameters which are treated as hyper-parameters. All the network weights are 
treated as proper model parameters. However, as our previous discussion shows,
in cases when there exists an environment of tasks posessing a common
internal representation, the hidden layer weights of a neural network 
should be viewed as hyper-parameters, not model parameters. The only model
parameters are the output weights. Thus, rather than the model parameters
vastly outnumbering the hyperparameters, we have the opposite situation 
here with the hyper-parameters vastly outnumbering the model parameters. 
We will see in the remainder of this paper that such an arrangement of
parameters is by far the most efficient, for two reasons. Firstly, once the 
hyperparameters have been learnt, \ie the objective prior has been identified,
then learning a novel task in the same environment requires only that the
model parameters be learnt, which for models with a small number of parameters
will be a relatively simple task and require few examples. Secondly, learning 
multiple tasks turns out be far more efficient when the hyper-parameters 
dominate the model parameters.

\section{Learning Multiple Tasks}
\label{results}
Having set up the model of Bayesian bias learning in the previous section, we
can now tackle the question posed in the introduction: 
``How much information is required per task to learn $n$ tasks simultaneously?''

Note that if the learner already knows the true prior 
$p(\theta|\pi^*)$, then the expected amount of information required 
per task to learn 
$n$ tasks is
\begin{equation}
\label{trueprior}
\frac{H(P_{\Theta^n|\pi^*})}n = H(P_{\Theta|\pi^*})
\end{equation}
because $P_{\Theta^n|\pi^*} = P_{\Theta|\pi^*}^n$ and entropy
is additive over products of independent distributions 
(here $H(P_{\Theta|\pi^*}) = 
-\E_{\Theta|\pi^*} \log p(\theta|\pi^*)$ is the entropy
of the true prior). As $H(P_{\Theta|\pi^*})$ is the expected amount of information
required to learn a single task, we can see that there is no advantage to 
learning multiple tasks if the true prior is known. 

If the true prior is unknown, but the learner is in posession of 
a family of priors $\Pi$, then the expected amount of information required 
per task to learn $n$ tasks is 
\begin{equation}
\label{info}
\Rbar_{n,\pi^*} := \frac{H_{\pi^*}(P_{\Theta^n})}{n},
\end{equation}
where $H_{\pi^*}(P_{\Theta^n}) 
= -\E_{\Theta^n|\pi^*} \log(p(\theta^n))$
where $p(\theta^n) = \int_\Pi p(\theta^n|\pi) p(\pi)\, d\pi$
is the {\em induced}
or {\em mixture} prior on $\theta^n$.
Rather than tackling $\Rbar_{n,\pi^*}$ directly it
is more convenient to analyse the expected difference between the 
information required to learn $n$ tasks using the true prior
$p(\theta^n|\pi^*)$ and the 
information required to learn $n$ tasks using the induced prior 
$p(\theta^n)$. This quantity is 
$$
\int_{\Theta^n} p(\theta^n|\pi^*) \log\frac{p(\theta^n|\pi^*)}{p(\theta^n)}
\, d\theta^n = D_K(P_{\Theta^n|\pi^*}\| P_{\Theta^n}),
$$
which is the {\em Kullback-Liebler divergence} between
the true and induced distributions on $\Theta^n$. 
Note that if we know $D_K(p_{\Theta^n|\pi^*}\| p_{\Theta^n})$
 we can recover $\Rbar_{n,\pi^*}$ from the relation
\begin{equation}
\label{rec}
\Rbar_{n,\pi^*}
= \frac1n D_K(P_{\Theta^n|\pi^*}\| P_{\Theta^n}) + H(P_{\Theta|\pi^*})
\end{equation}

To bound 
$D_K(P_{\Theta^n|\pi^*}\| P_{\Theta^n})$ the following definitions are needed.
\begin{defn}
\label{defs}
For any $\pi,\pi'\in\Pi$, let $\Delta_H(\pi,\pi')$ denote the 
squared Hellinger distance squared 
between the two priors $P_{\Theta|\pi}$ and $P_{\Theta|\pi'}$:
$$
\Delta_H(\pi,\pi') = \int_\Theta \[\sqrt{p(\theta|\pi)} - 
\sqrt{p(\theta|\pi')}\]^2\, d\theta
$$
and let $\Delta_K(\pi,\pi')$ denote the Kullback-Liebler divergence
between the two priors $p(\theta|\pi), p(\theta|\pi')$:
$$
\Delta_K(\pi,\pi') = \int_\Theta p(\theta|\pi) 
\log\frac{p(\theta|\pi)}{p(\theta | \pi')}\, d\theta.
$$
Let $B_\ep(\pi) = \{\pi'\colon \Delta^{1/2}_H(\pi,\pi') \leq \ep\}$, \ie
the Hellinger ball of radius $\ep$ around $\pi$.
For all $\pi\in \Pi$, define the {\em local metric dimension of $\pi$} by 
$$
\dim_{P_\Pi}(\pi) = \lim_{\ep\rightarrow 0}
\frac{- \log P_\Pi(B_\ep(\pi))}{\log\frac1\ep}
$$
whenever the limit exists ($P_\Pi$ is the subjective (hyper) prior probability 
distribution on $\Pi$).
\end{defn} 
Note that $(\Pi,\Delta^{1/2}_H)$ is a metric space while $(\Pi,\Delta_K)$ is not
($\Delta_K$ is asymmetric and does not satisfy the triangle inequality). 
Also, $\Delta_K(\pi,\pi') \geq \frac12\Delta_H(\pi,\pi')$ always (see 
\eg \cite{HausslerCOLT95}). 
\begin{defn}
Let $(X,\Sigma,P)$ be a measure space and $f,g\colon \N\times X\to \R$ be two
real-valued functions on $\N\times X$. We say 
$$
f(m,x) \thickapprox g(m,x) 
$$
if $\lim_{m\rightarrow\infty} P(X_m) = 1$ where for each $m\in\N$, 
$X_m = \{x\colon f(m,x) = g(m,x)\}$.
\end{defn}

\begin{thm}
\label{KLthm}
If there exists $\alpha < \infty$ such that for all $\pi,\pi'\in\Pi$,
$$
\Delta_K(\pi,\pi') \leq \alpha\Delta_H(\pi,\pi'),
$$
and $\dim_{P_\Pi}(\pi)$ exists for all $\pi\in\Pi$, then 
\begin{equation}
\label{limeqn}
\frac{D_K(P_{\Theta^n|\pi^*}\| P_{\Theta^n})}{\log n}
\thickapprox  \frac{\dim_{P_\Pi}(\pi^*)}{2} + o(1),
\end{equation}
where $o(1)$ is a function of $m$ for which $\lim_{m\rightarrow\infty} 
o(1)(m) = 0$.
\end{thm}
\begin{pf}
See section \ref{proofsec}.
\end{pf}
Note that if
$$
\inf_{\pi,\pi^*\in\Pi\,{\text{and}}\, \theta\in\Theta} 
\frac{p(\theta|\pi)}{p(\theta|\pi^*)} < \infty 
$$
then there exists $\alpha < \infty$ 
such that 
$\Delta_K(\pi,\pi') \leq \alpha\Delta_H(\pi,\pi')$ 
\cite{HausslerCOLT95}.

\begin{thm}
\label{easy}
Under the same conditions as theorem \ref{KLthm},
$$
\Rbar_{n,\pi^*} \thickapprox \frac{\dim_{P_\Pi}(\pi^*)}{2} \frac{\log n}{n} + 
H(P_{\Theta|\pi^*}) + o\(\frac{\log n}n\),
$$
where $o\(\log n/n\)$ approaches zero faster than $\log n/n$ 
as $n\rightarrow\infty$.
\end{thm}
\begin{pf}
The theorem follows directly from \eqref{rec} and theorem \ref{KLthm}.
\end{pf}
Note that this result is not quite as strong as it looks on face value because the
set of priors for which 
\begin{equation}
\label{fart}
\Rbar_{n,\pi^*} = \frac{\dim_{P_\Pi}(\pi^*)}{2} \frac{\log n}{n} + 
H(P_{\Theta|\pi^*}) + o\(\frac{\log n}n\)
\end{equation}
fails 
can {\em vary with $n$}, even though its measure becomes vanishingly small.
This implies that for any individual $\pi^*\in\Pi$, \eqref{fart} may fail 
for infinitely many $n$. 
However, if the sum over all $n$ of the $P_\Pi$ measure of the sets of
$\pi^*$ 
for which \eqref{fart} fails is finite, then by Borel-Cantelli, for all but a
set of $\pi$ of $P_\Pi$ measure zero, \eqref{fart} will fail only {\em
finitely} often.

Setting $a=H(P_{\Theta|\pi^*})$ and $b=\frac{\dim_{P_\Pi}(\pi^*)}{2}$, theorem \ref{easy} 
shows that the 
expected amount of information required per task to learn an
$n$ task training set approaches 
$$
a + \frac{b \log n}{n},
$$
except for a set of priors of vanishingly small measure as
$n\rightarrow\infty$,
which in turn approaches $a$---the minimum amount of information 
required to learn a task on average ($a$ is the amount of information 
required if the true prior is known, \cf \eqref{trueprior}).

\subsection{Example: learning an LDR}
Recall from section \ref{learnprior} that for the problem of learning 
a Low Dimensional Representation (LDR), $\Theta$ is split into 
$(\Theta_\LDR,\Theta_\OUT)$. We chose each prior $\pi\in\Pi$ to be delta function
over some $\theta_\LDR$ and uniform or Gaussian over $\Theta_\OUT$. In order
to apply the results of the previous section we need to smooth out the delta
functions, otherwise the correct prior is identifiable from the observation of 
a single task\footnote{We will put the delta function back in the next section
where we consider the more realistic scenario in which the learner receives 
information about $\theta$ in the form of examples $z$ chosen according 
to $p(z|\theta)$, rather than receiving $\theta$ directly.}
$\theta$. So instead take the prior for each $\pi$ to be a Gaussian with 
small variance $\sigma_\Pi$ peaked over some $\theta_\LDR$. In addition, for 
$H(P_{\Theta|\pi})$ to be well defined the output weights $\theta_\OUT$
need to be quantized, so let each weight $w$ be coded with $k$ bits and 
take the distribution over the discretized $\Theta_\OUT$ to be uniform for
each prior $\pi$. 
Denote the number of weights in $\Theta_\LDR$ by $W_\LDR$ and the number
of weights in $\Theta_\OUT$ by $W_\OUT$.
For any $\pi\in\Pi$, let $\theta_\LDR(\pi)$ denote the mean of the 
distribution $p(\theta_\LDR | \pi)$. Finally, take the prior 
distribution on $\Pi$ to be uniform over some compact subset 
of $\Theta_\LDR$.

A simple calculation shows the Hellinger and Kullback-Liebler 
distances to be given by
\begin{align*}
\Delta_H(\pi,\pi') &= 2\(1 - e^{\frac1{8\sigma_\Pi^2}\|\theta_\LDR(\pi) - 
\theta_\LDR(\pi')\|^2}\), \\
\Delta_K(\pi,\pi') &= \frac1{2\sigma_\Pi^2} 
\|\theta_\LDR(\pi) - \theta_\LDR(\pi')\|^2
\end{align*}
Note that as  $\Delta_H(\pi,\pi')\rightarrow 0$, 
$\Delta_H(\pi,\pi') \rightarrow 
\frac1{4\sigma_\Pi^2}\|\theta_\LDR(\pi) - \theta_\LDR(\pi')\|^2$. Substituting this
expression into the definition of $\dim_{P_\Pi}(\pi)$ we find
$$
\dim_{P_\Pi}(\pi) = W_{\LDR}
$$
for all $\pi\in\Pi$. 
Trivially, $H(P_{\Theta|\pi}) = k W_\OUT$ for all $\pi\in\Pi$.
The fact that the prior on $\Pi$ is compactly supported coupled with the use
of a Gaussian prior on $\Theta$ ensures that $\Delta_K(\pi,\pi')$ is bounded 
above by $\alpha \Delta_H(\pi,\pi')$ for all 
$\pi,\pi'$ and some $\alpha < \infty$. Hence the conditions of theorem
\ref{easy} are satisified and we have
$$
\Rbar_{n,\pi^*} \thickapprox \frac{W_\LDR}{2} \frac{\log n}{n} +  
k W_\OUT + o\(\frac{\log n}n\).
$$
The similarity of this expression to the upper bound on the number
of examples required per task for good generalisation in a PAC
sense of $O(W_\OUT + W_\LDR/n)$ is noteworthy (see \cite{colt95} for
a derivation of the latter expression).

\input{help}

\input{proof}
}
\bibliographystyle{abbrv}
\bibliography{bib}
\end{document}

%% file: mathdefs.tex
\newcommand{\bb}{\mathbb}
\renewcommand{\Box}{\qedhere}
\newcommand{\mathbold}[1]{\mbox{\boldmath $\bf#1$}}

\renewcommand{\glossary}[2]{}
\def\addcontentsline#1#2#3{} 
\renewcommand{\thickapprox}{\doteq}
\newtheorem{thm}{Theorem}

\newtheorem{lem}[thm]{Lemma}

\newtheorem{corr}[thm]{Corollary}
\newtheorem{defn}{Definition}

\newcommand{\ie}{{\em i.e.\ }}

\newcommand{\iid}{{i.i.d.\ }}

\newcommand{\eg}{{\em e.g.\ }}
\newcommand{\cf}{{\em c.f.\ }}
\newcommand{\etc}{{\em etc\ }}

\newcommand{\E}{\bb E}

\newcommand{\Zn}{{Z^n}}

\newcommand{\Znm}{{Z^{(n,m)}}}
\newcommand{\Znmp}{{Z^{(n,m+1)}}}

\newcommand{\znm}{{z^{(n,m)}}}

\newcommand{\z}{{\mathbold{z}}}

\newcommand{\be}{\begin{equation*}}

\newcommand{\ep}{{\varepsilon}}

\newcommand{\N}{{\cal N}}

\renewcommand{\(}{\left(}
\renewcommand{\)}{\right)}
\renewcommand{\[}{\left[}
\renewcommand{\]}{\right]}

\newcommand{\Rbar}{{\overline R}}

\renewcommand{\hbar}{{\overline h}}

\newcommand{\thetat}{{\tilde{\theta}}}

\newcommand{\LDR}{{\text{\rm LDR}}}
\newcommand{\OUT}{{\text{\rm OUT}}}

\newcommand{\R}{{\mathbb R}}

%% file: help.tex
\section{Sampling multiple tasks}
\label{nmsec}
Theorem \ref{easy} was derived under the assumption that the learner
receives information about the tasks $\theta$ directly. In fact $\Rbar_{n,\pi^*}$ 
is (within one query) 
the average number of {\em queries} the learner will require to identify 
a task in an $n$-task training set if the queries are restricted to be of
the form ``is $\theta\in A$'' where $A$ is any subset of $\Theta$ and the learner
uses the best possible querying strategy.

In general the learner will not be able to query in this way, but instead
will receive information about the parameters $\theta$ indirectly via a sample 
$z^m = (z_1,\dots,z_m)$, sampled  \iid according to  $p(z|\theta)$. If the 
learner is learning $n$ tasks simultaneously then it will recieve 
$n$ such samples
(called an {\em (n,m)-sample} in 
\cite{colt95,BiasLearn95}):
$$
\z =
\begin{matrix}
z_{11} & \hdots & z_{1m} \\
\vdots & \ddots & \vdots \\
z_{n1} & \hdots & z_{nm} 
\end{matrix}
$$
Let $\Znm$ denote the set of all such $\znm$\footnote{Note that conditional
upon our prior knowledge (represented by the hyper-prior $P_\Pi$), 
the columns of $\z$ are {\em independently} 
distributed, while the rows of $\z$ are {\em identically} distributed. 
Thus $\Znm$ is in some sense the simplest, non-trivial 
(not all entries \iid) {\em matrix} of random variables.}.
The correct hierarchical Bayes approach 
to learning the $n$ tasks $\theta_1,
\dots,\theta_n$ is to use the 
hyper prior $P_\Pi$ to generate a prior distribution on $\Theta^n$ via
\begin{align*}
p(\theta^n) &= \int_\Pi p(\theta^n|\pi) p(\pi)\, d\pi \\
            &= \int_\Pi p(\pi) \prod_{i=1}^n p(\theta_i|\pi)\, d\pi 
\end{align*}
and then the posterior $p(\theta^n|\z)$ can be computed in the usual 
way
\begin{align}
\label{blurb}
p(\theta^n|\z) &= \frac{p(\z|\theta^n) p(\theta^n)}{p(\z)} \\*
&= \frac{p(\theta^n) \prod_{i=1}^n \prod_{j=1}^m p(z_{ij}|\theta_i)}{p(\z)}
\end{align}
where $p(\z) = \int_{\Theta^n} p(\theta^n) 
\prod_{i=1}^n \prod_{j=1}^m p(z_{ij}|\theta_i) \, d\theta^n$.

One way to measure the advantage in learning $n$ tasks together is by the rate
at which the learner's loss in predicting novel examples decays for each task. In
keeping with our philosophy of measuring loss in information terms
(\ie via relative entropy), the expected loss per task of the learner when 
predicting the $m+1$th observation of each task, $z^n_{m+1}$, 
after receiving $\z$, is
\begin{equation}
\label{nm+1loss}
\Rbar_{n,m,\pi^*} = \frac1n\E_{\Theta^n|\pi^*}\E_{\Znm|\theta^n}\E_{\Zn|\theta^n}
\log\frac{p(z^n|\theta^n)}{p(z^n|\z)},
\end{equation}
where $p(z^n|\z)$ is the learner's predictive distribution on $\Zn$
based on the information contained in $\z$ and is given by
$$
p(z^n|\z) = \int_{\Theta^n} p(z^n|\theta^n) p(\theta^n|\znm) \, d\theta^n.
$$
where $ p(\theta^n|\znm)$ is computed via \eqref{blurb}.
Note that \eqref{nm+1loss} is also the expected loss of a learner that has
first received $m$ observations of $n$ tasks, $\z$, 
then observed $m$ observations of a new task, and is predicting the $m+1$th
observation the new task. In this way it is a measure of the extent to which 
the learner has {\em learnt to learn} the tasks in the environment after 
receiving $\z$.

Let $p_\Znm$ denote the learner's prior distribution on $\Znm$ induced
by $P_{\Theta^n}$ (which in turn is induced by $P_\Pi$):
\begin{align*}
p_\Znm(\znm) &= \int_{\Theta^n} p(\znm|\theta^n) p(\theta^n)\,d\theta^n \\
&= \int_\Pi \int_{\Theta^n} p(\znm|\theta^n) p(\theta^n|\pi) p(\pi) \,d\theta^n
\, d\pi.
\end{align*}
For any $\theta^n\in\Theta^n$, define $\dim_{P_{\Theta^n}}(\theta^n)$ 
as in definition \ref{defs}:
$$
\dim_{P_{\Theta^n}}(\theta^n) = \lim_{\ep\rightarrow 0} 
\frac{-\log P_{\Theta^n} \(B_\ep(\theta^n)\)}{\log\frac1\ep}
$$
whenever the limit exists. Define $\Delta_H(\theta^n,\thetat^n)$ and 
$\Delta_K(\theta^n,\thetat^n)$ respectively as the Hellinger and KL divergences 
between the distributions induced on $\Zn$ by $\theta^n$ and $\thetat^n$
(as in definition \ref{defs}).
\begin{thm}
\label{gthm}
For this theorem fix $n\in N$ 
and take all limiting behaviour to be with respect to $m$.
Assume there exists $\alpha < \infty$ such that for all 
$\theta,\theta'\in\Theta$,
$$
\Delta_K(\theta,\theta') \leq \alpha\Delta_H(\theta,\theta'),
$$
and that $\dim_{P_{\Theta^n}}(\theta^n)$ exists for all 
$\theta^n$ such that 
$p(\theta^n|\pi^*) > 0$. Suppose also that 
$P_{\Theta^n | \pi^*}$ is absolutely continuous with respect to 
$P_{\Theta^n}$. Finally, assume 
$m \Rbar_{n,m,\pi^*} = d + o(1)$ for some $d$.
Then,
$$
\Rbar_{n,m,\pi^*}  =
\frac1{2nm}\E_{\Theta^n|\pi^*}\dim_{P_{\Theta^n}}(\theta^n) + o\(\frac1m\).
$$
\end{thm}
\begin{pf}
One can easily verify that 
\begin{multline}
\label{stuffed}
\Rbar_{n,m,\pi^*} = \frac1n\E_{\Theta^n|\pi^*}  
\(D_K(P_{\Znmp|\theta^n}\| P_\Znmp)\right.\\
 \left.- D_K(P_{\Znm|\theta^n}\|P_\Znm)\)
\end{multline}
As $\Delta_K(\theta^n,\thetat^n) = \sum_{i=1}^n \Delta_K(\theta_i,\thetat_i)$,
the condition $\Delta_K(\theta,\theta') \leq \alpha \Delta_H(\theta,\theta')$ 
ensures the same condition holds for $\Delta_K(\theta^n,\thetat^n)$ with $\alpha$
replaced by $n\alpha$. By the definition of $\Rbar_{n,m,\pi^*}$ we only need
to consider those $\theta^n$ for which $p_{\Theta^n|\pi^*}(\theta^n) > 0$, 
and we have 
assumed that $\dim_{P_{\Theta^n}}(\theta^n)$ exists for those values. So we 
can apply theorem \ref{KLthm} (with $\Pi$ replaced by $\Theta^n$  and $n$ replaced 
by $m$) to the $D_K's$ in the right-hand-side of \eqref{stuffed}. This gives 
\begin{multline}
\label{hell}
\Rbar_{n,m,\pi^*} \thickapprox \frac1{2n}\(\log(m+1) - \log m\)
\E_{\Theta^n|\pi^*}\dim_{P_\Theta^n}(\theta^n)\\ + o(\log(m+1)) - o(\log m). 
\end{multline}
Note the absolute continuity condition is needed to ensure that 
the measure of the 
set of $\theta^n$ failing the equality $D_K(P_{\Znm|\theta^n}\| P_\Znm) = 
1/2\dim_{P_{\Theta^n}}(\theta^n)\log m + o(\log m)$ has $P_{\Theta^n|\pi^*}$
measure zero in the limit of large $m$, as well as $P_{\Theta^n}$ measure zero.

Without the $o(\log(m+1)) - o(\log m)$ term in \eqref{hell} the result would
be immediate, as $\log(m+1) - \log m \rightarrow 1/m$. However, the assumption
$m \Rbar_{n,m,\pi^*} \thickapprox d + o(1)$ for some $d$ is needed to ensure
that $o(\log(m+1)) - o(\log m) = o(1/m)$. To show this we need 
the following lemma:
\begin{lem}
\label{fartybreath}
Suppose $a,b\colon \N\times X\to \R$ are such that 
$a(m,x) = \sum_{k=1}^m b(k,x)$ for all $x\in X$. Suppose also that 
$a(m,x)/\log m \thickapprox d + o(1)$. If 
$m b(m,x) \thickapprox d' + o(1)$, then $d'=d$.
\end{lem}
\begin{pf}
By the assumptions of the lemma, $b(m,x) \thickapprox d'/m + o(1/m)$ which 
means there exists $h(m)$ such that $m h(m) \rightarrow 0$ and the sets
$X_m = \{x\colon d'/m - h(m) \leq b(m,x) \leq d'/m + h(m)\}$ satisfy 
$P(X_m) \rightarrow 1$. Fix $x\in X_m$
As $a(m,x) = \sum_{k=1}^m b(k,x)$, 
\begin{equation}
\label{temp}
\sum_{k=1}^m \frac{d'}{k} - \sum_{k=1}^m h(k) \leq a(m,x) \leq \sum_{k=1}^m
\frac{d'}{k} + \sum_{k=1}^m h(k).
\end{equation}
Now, there exists a constant $c$ such that $\left|\sum_{k=1}^m 1/k -\log
m\right| \leq c$ for all $m$, and so $d'\log m - d' c \leq \sum_{k=1}^m d'/k
\leq d'\log m + d' c$. Let $h_a(m) = \sum_{k=1}^m h(k)$. As $m h(m)
\rightarrow 0$, we can apply lemma 6 from \cite{HausslerCOLT95} to get 
$h_a(m)/\log m \rightarrow 0$. Substituting into \eqref{temp} yields,
$$
d' \log m - d'c - h_a(m) \leq a(m,x) \leq d'\log m + d'c + h_a(m)
$$
for all $x\in X_m$. As $\frac{h_a(m) + d'c}{\log m} \rightarrow 0$, we have 
shown that $a(m,x) \thickapprox d'\log m + o(\log m)$, as required.
\end{pf}

Define 
$$
\Rbar_{n,0,\pi^*} = \frac1n \E_{\Theta^n|\pi^*} \E_{\Zn|\theta^n} 
\log \frac{p(z^n|\theta^n)}{p(z^n)}.
$$
Explicit calculation shows,
$$
\sum_{k=0}^m \Rbar_{n,k,\pi^*} = \frac1n \E_{\Theta^n|\pi^*} 
D_K\(P_{\Znmp|\theta^n}\| P_{\Znmp}\).
$$
By theorem \ref{KLthm} again we know that 
$$
\frac{D_K\(P_{\Znmp|\theta^n}\| P_{\Znmp}\)}{\log m} 
\thickapprox \frac{\dim_{P_{\Theta^n}}(\theta^n)}{2} + o(1),
$$
for all $\theta^n$ such that $p(\theta^n|\pi^*) > 0$.
Hence, applying lemma \ref{fartybreath} and taking expectations we have that 
$$
\Rbar_{n,m,\pi^*} = 
\frac1{2mn}\E_{\Theta^n|\pi^*}\dim_{P_{\Theta^n}}(\theta^n) + o\(\frac1m\)
$$
as required.
\end{pf}

In the course of proving theorem \ref{gthm} we have also proved the following 
corollary bounding the average {\em cumulative} loss of the learner:
\begin{corr}
\label{dumbbum}
Under the same conditions as theorem \ref{gthm} (except that the
condition $m \Rbar_{n,m,\pi^*} = d + o(1)$ for some $d$ is
not necessary),
$$
\sum_{k=0}^m \Rbar_{n,k,\pi^*} =
\frac{\log m}{2 n}\E_{\Theta^n|\pi^*} \dim_{P_{\Theta^n}}(\theta^n)
+ o(\log m).
$$
\end{corr}

Theorem \ref{gthm} and corrrolary \ref{dumbbum} give expressions for
the asymptotic average {\em instantaneous} loss and average 
asymptotic {\em cumulative} loss for a learner that is simultaneously
learning $n$ tasks using a hierarchical model. If the learner does not 
take account of the fact that the $n$ tasks are related then each time it
comes to learn a new task it will start with the same prior 
$p(\theta) = \int_\Pi p(\theta|\pi) p(\pi) \, d\pi$. Thus, using theorem
\ref{gthm} with $n=1$, the learner's average 
instantaneous loss when learning $n$ tasks will in this case be given by
$$
\frac{1}{2m}\E_{\Theta|\pi^*} \dim_{P_\Theta}(\theta) + o(\frac1m),
$$
while corollary \ref{dumbbum} shows that the average cumulative loss 
of the learner will be
$$
\frac{\log m}{2}\E_{\Theta|\pi^*} \dim_{P_\Theta}(\theta) + o(\log m).
$$
Thus the difference between the learner's loss when taking task relatedness
into account vs. ignoring task relatedness is captured by the 
difference between
\begin{equation}
\label{n}
\frac1n\E_{\Theta^n|\pi^*} \dim_{P_{\Theta^n}}(\theta^n)
\end{equation}
and
\begin{equation}
\label{one}
\E_{\Theta|\pi^*} \dim_{P_{\Theta}}(\theta).
\end{equation}
In the next section we calculate expressions \eqref{n} and \eqref{one} 
for a general class of hierarchical models that includes the LDR model.

\subsection{Dimension of Smooth Euclidean Hierarchical Models}
\label{dimsec}
We now specialise to the case where $\Pi = \R^b$, $\Theta = \R^a\times\R^b$
and 
\begin{equation}
\label{exam}
p(\theta = (x^a,x^b) | \pi) = \delta(x^b - \pi) g_\pi(x_a) 
\end{equation}
where $\delta(\cdot)$ is the $b$-dimensional Dirac delta function and 
$g_\pi$ is a twice differentiable function on $\R^a$. Let $p(\pi) = f(\pi)$ 
where $f$ is also twice-differentiable. This model includes 
the LDR model discussed in section \ref{learnprior},
$(\theta_\OUT,\theta_\LDR) = (x^a,x^b)$, 
as well as any smooth model in which there are 
$a+b$ real parameters, $b$ of which are effectively hyperparameters and are 
fixed by the prior and the remainding $a$ of which are model parameters. 
This hierarchical model will be referred to as an $a:b$ model. 
\begin{defn}
Let $(X,\rho)$ be a metric space. We say a second metric $\rho'$ 
{\em locally dominates} $\rho$ if for all $x\in X$, there exists
$\ep,c,c' >0$ such that for all $x'\in B_\ep(x,\rho)$ (the $\ep$-ball around
$x$ under $\rho$),
$$
c\rho'(x,x') \leq \rho(x,x') \leq c' \rho'(x,x').
$$
\end{defn}

\begin{thm}
\label{dimthm}
Let $\Pi,\Theta,Z$ and $p(\pi),p(\theta|\pi)$ define an $a:b$ model
and suppose the conditional distributions $p(z|\theta)$ are 
such that $\Delta_H^{1/2}$ is locally dominated 
by $\|\cdot\|$ on $\Theta$. For all $\theta^n$ such that
$p(\theta^n) = \E_\Pi p(\theta^n|\pi) > 0$,
$$
\dim_{P_{\Theta^n}}(\theta^n) = n a + b.
$$
In addition, for any $\pi$ and for all $\theta^n$ such that $p(\theta^n|\pi)
> 0$,
$$
\dim_{P_{\Theta^n|\pi}}(\theta^n) = n a.
$$
\end{thm}
\begin{pf}
Omitted.
\end{pf}
So for $a:b$ models in which $\Delta_H^{1/2}$ is locally dominated 
by $\|\cdot\|$, expressions \eqref{n} and \eqref{one} reduce to
\begin{align*}
\frac1n\E_{\Theta^n|\pi^*} \dim_{P_{\Theta^n}}(\theta^n)
&= a + \frac{b}{n}, \\
\E_{\Theta|\pi^*} \dim_{P_{\Theta}}(\theta) &= a + b.
\end{align*}
Hence the learner's average instantaneous loss when learning $n$ tasks 
will be 
\begin{equation}
\label{burp}
\Rbar_{n,m,\pi^*} = \frac{1}{2m}\(a + \frac{b}{n}\) + o\(\frac1m\)
\end{equation}
if the tasks are learnt hierarchically, and
$$
\Rbar_{n,m,\pi^*} = \frac{1}{2m}\(a + b\) + o\(\frac1m\)
$$
if they are learnt independently. Similar expressions hold for the cumulative 
loss. Thus, the hierarcichal approach always does better asymptotically, and 
is most advantageous when the hyperparameters dominate the parameters 
($b >> a$). In addition, if the true prior is known then application of the
second part of theorem \ref{dimthm} shows that asymptotically the
instantaneous risk satisfies
\begin{equation}
\label{burp1}
\Rbar_{n,m,\pi^*} = \frac{a}{2m} + o\(\frac1m\),
\end{equation}
with a similar expression for the cumulative risk.
Comparing \eqref{burp} with \eqref{burp1}, we see that the effect of lack of
knowledge of the true prior can be made arbitrarily small by learning enough
tasks simultaneously.

The following theorem, proof omitted,
gives two conditions under which $\|\cdot\|$
locally dominates $\Delta_H^{1/2}$. 
\begin{thm}
\label{barf}
If the map $P_{Z|\theta} \mapsto \theta$ is continuous 
(\ie $P_{Z|\theta}\rightarrow P_{Z\theta_0} \Rightarrow
\theta\rightarrow\theta_0$ where convergence on the left is weak convergence)
and the Fisher information matrix  
$$
J(\theta) = \E_{Z|\theta}\left[\frac{\partial}{\partial\theta_i}
\log p(z|\theta) \frac{\partial}{\partial\theta_j}
\log p(z|\theta)\right]_{i,j=1,\dots,a+b}
$$
exists and is positive definite for all $\theta$ then 
$\Delta_H^{1/2}$ is locally dominated 
by $\|\cdot\|$ on $\Theta$. 
\end{thm}
For the LDR neural network model, $p(y=1,x|\theta) = p(x)f_\theta(x)$,
the condition $P_{Z|\theta} \mapsto \theta$ is continuous fails because 
the network is invariant under the group of transformations consisting of
hidden-layer node permutations and sign-changes of all incoming and 
outgoing weights at each node. However, it is known that these are the only
symmetry transformations of the class of one-hidden-layer, sigmoidal nets with
linear output nodes (see \cite{sussman,sontagseig,kurkova}). So if we 
work in the ``factor'' space of networks in which all these permutations and
sign changes of a weight vector are identified, the continuity condition
will be satisfied. Hence with a little more work we can prove the following:
\begin{thm}
\label{LDR}
$\Delta_H^{1/2}(\theta,\theta')$ is locally dominated 
by $\|\theta-\theta'\|$ for the single-hidden layer, linear-output 
LDR model.
\end{thm}
In this case $a=W_\OUT$ and $b=W_\LDR$ where $W_\OUT$ is the number 
of weights in an output node and $W_\LDR$ are the number of input-hidden 
weights. Hence,
\begin{equation}
\label{burp3}
\Rbar_{n,m,\pi^*} = \frac{1}{2m}\(W_\OUT + 
\frac{W_\LDR}{n}\) + o\(\frac1m\).
\end{equation}
Again the advantage in learning multiple tasks when the true prior is unknown
is clear, and
parallels precisely the upper bounds of the VC/PAC model (recall equation
\eqref{eq1}).

\section{Conclusion}
The problem of learning appropriate domain-specific bias via multi-task 
sampling has been modelled 
from a Bayesian/Information theoretic viewpoint. The approach shows that 
in many high-dimensional, essentially ``non-parametric'' modelling scenarios,
most of the model parameters are more appropriately regarded as
hyper-parameters. Performing hierarchical Bayesian inference within such a
model, using multiple task sampling, is asymptotically much more efficient
than a non-hierarchical approach. 

An interesting avenue for further investigation would be to examine the 
asymptotic (as a function of $n$ and $m$) 
behaviour of the posterior distribution on the space of priors,
$p(\pi|\z)$. This would extend known results on the asymptotic normality 
of the posterior in ordinary, parametric Bayesian inference (see \eg
\cite{CB1}).

%% file: proof.tex
\section{Proof of theorem \ref{KLthm}}
\label{proofsec}
Let $I(\Pi;\Theta^n)$ denote the {\em mutual information} between $\Pi$ and
$\Theta^n$, which can easily be seen to satisfy
$$
I(\Pi,\Theta^n) = \E_{\Pi^*} D_K(P_{\Theta^n|\pi^*}\| P_{\Theta^n}).
$$
The following theorem is theorem 1 from \cite{HausslerCOLT95}.
\begin{thm}[\cite{HausslerCOLT95}]
\label{ithm}
For all $n\geq 1$,
\begin{align*}
-\E_{\Pi^*} \log \E_\Pi e^{- \frac{n}{4}\Delta_H(\pi^*,\pi)}
&\leq I(\Pi,\Theta^n)\\  
&= \E_{\Pi^*} D_K(P_{\Theta^n|\pi^*}\| P_{\Theta^n})\\
&\leq -\E_{\Pi^*} \log \E_\Pi e^{-n\Delta_K(\pi,\pi^*)}
\end{align*}
\end{thm}
Using the assumption of the theorem that 
$\Delta_K(\pi,\pi') \leq \alpha \Delta_H(\pi,\pi')$ we 
have:
\begin{align}
\label{hello}
-\E_{\Pi^*} \log \E_\Pi e^{- \frac{n}{4}\Delta_H(\pi^*,\pi)}
&\leq I(\Pi,\Theta^n) \\
&\leq -\E_{\Pi^*} \log \E_\Pi e^{-n\alpha\Delta_H(\pi,\pi^*)}
\end{align}

For any pair of random variables $W$ and $V$ and any real-valued
function $u(w,v)$, we have the following inequality due to Feynman:
\begin{equation}
\label{fineq}
- \E_V \log \E_W e^{u(w,v)} \leq -\log \E_W e^{\E_V u(w,v)}.
\end{equation}
Using \eqref{fineq} we can effectively ``lop off'' the expectation over $\Pi^*$ 
in the upper bound of \eqref{hello} to give an upper bound on
$D_K(P_{\Theta^n|\pi^*}\| P_{\Theta^n})$.
\begin{lem}
\label{udk}
For all $n\geq 1$ and $\pi^*\in\Pi$,
$$
D_K(P_{\Theta^n|\pi^*}\| P_{\Theta^n}) \leq -\log \E_\Pi  e^{-n\alpha\Delta_H(\pi,\pi^*)}
$$
\end{lem}
\begin{pf}
The proof is via the same chain of inequalities used to prove 
the upperbound in theorem \ref{ithm}.
\begin{align*}
D_K(P_{\Theta^n|\pi^*}\| P_{\Theta^n}) &= \E_{\Theta^n|\pi^*} 
\log\frac{p_{\pi^*}(\theta^n)}{\E_\Pi p_\pi(\theta^n)} \\
&= - \E_{\Theta^n|\pi^*}\log\E_\Pi e^{\log\frac{p_\pi(\theta^n)}
{p_{\pi^*}(\theta^n)}} \\
&\leq -\log \E_\Pi e^{\E_{\Theta^n|\pi^*}\log\frac{p_\pi(\theta^n)}
{p_{\pi^*}(\theta^n)}} \\
&= -\log E_\Pi e^{- D_K(P_{\Theta^n|\pi^*}\| P_{\Theta^n | \pi})} \\  
&= -\log\E_\Pi e^{-n \Delta_K(\pi,\pi^*)} \\
&\leq -\log \E_\Pi  e^{-n\alpha\Delta_H(\pi,\pi^*)}.
\end{align*}
The penultimate inequality follows because the KL divergence is additive over the product
of independent distributions (see \eg \cite{CT}), and the last inequality follows
from the assumptions of the theorem.
\end{pf}

\begin{lem}
\label{dimlem}
If $\dim_{P_\Pi}(\pi^*)$ exists then
for any $0 < \alpha < \infty$,
$$
\lim_{n\rightarrow\infty}
\frac{-\log \E_\Pi e^{-n\alpha\Delta_H(\pi,\pi^*)}}{\log n} =
\frac{\dim_{P_\Pi}(\pi^*)}{2}. 
$$
\end{lem}
\begin{pf}
The arguments used in this proof are similar to those used in 
\cite{HausslerCOLT95} for proving corresponding global metric entropy 
bounds.

Setting $\ep = \frac1{\sqrt{\alpha n}}$, we have
$$
\frac{-\log \E_\Pi e^{-n\alpha\Delta_H(\pi,\pi^*)}}{\log n} 
= \frac{-\log \E_\Pi e^{-\(\frac1\ep\Delta_H^{1/2}(\pi,\pi*)\)^2}}
{-2\log \ep -\log \alpha}.
$$
Set $\ep$ sufficiently small to ensure that $-2\log \ep -\log\alpha > 0$.
Now,
\begin{align*}
- \log \E_\Pi &e^{-\(\frac1\ep\Delta_H^{1/2}(\pi,\pi*)\)^2}  \\
&= 
- \log\(\int_{B_\ep(\pi^*)} p(\pi)  
e^{-\(\frac1\ep\Delta_H^{1/2}(\pi,\pi*)\)^2}\, d\pi \right. \\
&\qquad + \left.\int_{B^c_\ep(\pi^*)} p(\pi)  
e^{-\(\frac1\ep\Delta_H^{1/2}(\pi,\pi*)\)^2}\, d\pi \) \\
&\leq - \log\[\frac1e p(B_\ep(\pi^*))\] \\
&= - \log p\(B_\ep(\pi^*)\) + 1,
\end{align*}
and so 
\begin{align*}
\limsup_{\ep\rightarrow 0} 
&\frac{-\log \E_\Pi e^{-\(\frac1\ep\Delta_H^{1/2}(\pi,\pi*)\)^2}}
{-2\log \ep - \log \alpha} \\ 
&\leq \limsup_{\ep\rightarrow 0}  
\frac{- \log p(B_\ep(\pi^*)) + 1}{- 2\log \ep - \log \alpha} \\
&= \frac{\dim_{P_\Pi}(\pi^*)}{2}.
\end{align*}
To get a matching lower bound note that 
for all $r > 0$,
\begin{align*}
- \log \E_\Pi &e^{-\(\frac1\ep\Delta_H^{1/2}(\pi,\pi*)\)^2} \\
&= 
- \log\(\int_{B_r(\pi^*)} p(\pi)  
e^{-\(\frac1\ep\Delta_H^{1/2}(\pi,\pi*)\)^2}\, d\pi \right. \\
&\qquad + \left.\int_{B^c_r(\pi^*)} p(\pi)  
e^{-\(\frac1\ep\Delta_H^{1/2}(\pi,\pi*)\)^2}\, d\pi \) \\
&\geq \log\[p\(B_r(\pi^*)\) + e^{-\(\frac{r}{\ep}\)^2}\].
\end{align*}
Setting $r = \ep^{1-\delta}$ gives
$$
-\log \E_\Pi e^{-\(\frac1\ep\Delta_H^{1/2}(\pi,\pi*)\)^2} \geq 
-\log\(p\(B_{\ep^{1-\delta}}(\pi^*)\) + e^{-\frac{1}{{\ep}^{2\delta}}}\)
$$
Because $\dim_{P_\Pi}(\pi^*)$ exists, we know that $p\(B_{\ep^{1-\delta}}(\pi^*)\)$ decreases
no faster than some power of $\ep^{1-\delta}$. However, for all $\delta > 0$, 
$e^{-\frac{1}{{\ep}^{2\delta}}}$ decreases faster than any polynomial as 
$\ep\rightarrow 0$. Thus
$$
\lim_{\ep\rightarrow 0} 
\frac{-\log\(p\(B_{\ep^{1-\delta}}(\pi^*)\) + e^{-\frac{1}{{\ep}^{2\delta}}}\)}
{-\log\ep} = (1-\delta)\dim_{P_\Pi}(\pi^*)
$$
for all $\delta > 0$, and so  
$$
\liminf_{\ep\rightarrow 0} 
\frac{-\log \E_\Pi e^{-\(\frac1\ep\Delta_H^{1/2}(\pi,\pi*)\)^2}}
{-2\log \ep - \log \alpha} \geq \frac{1-\delta}{2}\dim_{P_\Pi}(\pi^*)
$$
for all $\delta > 0$. Letting $\delta \rightarrow 0$ finishes the proof. 
\end{pf}

From lemmas \ref{dimlem} and \ref{udk},
\begin{equation}
\label{dup}
\limsup_{n\rightarrow \infty} 
\frac{D_K(P_{\Theta^n|\pi^*}\| P_{\Theta^n})}{\log n} \leq \frac{\dim_{P_\Pi}(\pi^*)}{2}.
\end{equation}
Applying lemma \ref{dimlem} to theorem \ref{ithm} and invoking Fatou's 
lemma twice gives
\begin{equation}
\label{ilim}
\lim_{n\rightarrow\infty} 
\frac{\E_{\Pi^*} D_K(P_{\Theta^n|\pi^*}\| P_{\Theta^n})}{\log n} 
= \E_{\Pi^*} \frac{\dim_{P_\Pi}(\pi^*)}{2}.
\end{equation}
Now let 
\begin{multline*}
\Pi_{\text{supbad}} = \left\{\pi^*\in\Pi\colon 
\limsup_{n\rightarrow\infty} D_K(P_{\Theta^n|\pi^*}\| P_{\Theta^n}) \right.\\
\left.< \frac{\dim_{P_\Pi}(\pi^*)}{2}\right\}
\end{multline*}
Suppose that $P_\Pi(\Pi_{\text{supbad}}) > 0$.
Then,
\begin{align*}
\E_{\Pi} \frac{\dim_{P_\Pi}(\pi)}{2}
&= \limsup_{n\rightarrow\infty} 
\E_\Pi \frac{D_K(P_{\Theta^n|\pi}\| P_{\Theta^n})}{\log n} \\
&\leq \limsup_{n\rightarrow\infty}\E_{\Pi_{\text{supbad}}} 
\frac{D_K(P_{\Theta^n|\pi}\| P_{\Theta^n})}{\log n} \\
&\qquad+ \limsup_{n\rightarrow\infty}\E_{\Pi^c_{\text{supbad}}} 
\frac{D_K(P_{\Theta^n|\pi}\| P_{\Theta^n})}{\log n}\\
&\leq \E_{\Pi_{\text{supbad}}} 
\limsup_{n\rightarrow\infty}\frac{D_K(P_{\Theta^n|\pi}\| P_{\Theta^n})}{\log n} \\
&\qquad+ \E_{\Pi^c_{\text{supbad}}} 
\limsup_{n\rightarrow\infty}\frac{D_K(P_{\Theta^n|\pi}\| P_{\Theta^n})}{\log n}\\
&< \E_{\Pi_{\text{supbad}}} \frac{\dim_{P_\Pi}(\pi)}{2}
+ \E_{\Pi^c_{\text{supbad}}} \frac{\dim_{P_\Pi}(\pi)}{2} \\
&= \E_\Pi \frac{\dim_{P_\Pi}(\pi)}{2},
\end{align*}
a contradiction. Thus $P_\Pi(\Pi_{\text{supbad}}) = 0$. 
Now, for each $n=1,2,\dots$ and 
$\ep > 0$ let 
$$
\Pi_{n,\ep} = \{\pi\colon\frac{D_K(P_{\Theta^n|\pi}\| P_{\Theta^n})}{\log n} 
< \frac{\dim_{P_\Pi}(\pi)}{2} - \ep\}.
$$ 
Suppose that 
$\limsup_{n\rightarrow\infty} P_{\Pi}\(\Pi_{n,\ep}\) = \kappa > 0$. Hence there 
exists an infinite sequence of integers $n_1 < n_2 <\dots$ such that 
$P_\Pi\(\Pi_{n_i,\ep}\) \geq \kappa$. From \eqref{dup} we know that for any 
$0 < \delta < \ep \kappa$ there exists $k > 0$ such that for all $i> k$,
$$
\frac{D_K(P_{\Theta^{n_i}|\pi}\| P_{\Theta^{n_i})}}{\log n_i} < 
\frac{\dim_{P_\Pi}(\pi)}{2} + \ep\kappa - \delta.
$$
Hence, for all $i\geq k$,
\begin{align*}
\E_\Pi &\frac{D_K(P_{\Theta^{n_i}|\pi}\| P_{\Theta^{n_i}})}{\log {n_i}} \\
&= \E_{\Pi_{n_i,\ep}} \frac{D_K(P_{\Theta^{n_i}|\pi}\| P_{\Theta^{n_i}})}{\log {n_i}}
+\E_{\Pi^c_{n_i,\ep}} 
\frac{D_K(P_{\Theta^{n_i}|\pi}\| P_{\Theta^{n_i}})}{\log {n_i}} \\
&<\E_{\Pi_{n_i,\ep}} \frac{\dim_{P_\Pi}(\pi)}{2} - \ep\kappa
+\E_{\Pi^c_{n_i,\ep}} \frac{\dim_{P_\Pi}(\pi)}{2} + \ep\kappa -\delta \\
&= \E_\Pi \frac{\dim_{P_\Pi}(\pi)}2 - \delta.
\end{align*}
and so 
\begin{align*}
\E_\Pi\frac{\dim_{P_\Pi}(\pi)}{2} 
&=
\lim_{i\rightarrow\infty} 
\E_\Pi \frac{D_K(P_{\Theta^{n_i}|\pi}\| P_{\Theta^{n_i}})}{\log {n_i}}  \\
&\leq \E_\Pi\frac{\dim_{P_\Pi}(\pi)}{2} - \delta,
\end{align*}
which is a contradiction and so the assumption
$\limsup_{n\rightarrow\infty} P_{\Pi}\(\Pi_{n,\ep}\)  > 0$ must be false. 
Hence for all $\ep >0$, $\lim_{n\rightarrow\infty} P_{\Pi}\(\Pi_{n,\ep}\)  =
0$. Setting  
\begin{multline*}
\Pi'_{n,\ep} = \left\{\pi\colon\frac{D_K(P_{\Theta^n|\pi}\| P_{\Theta^n})}{\log n} 
< \frac{\dim_{P_\Pi}(\pi)}{2} - \ep  \right.\\
\left.\text{or}\quad 
\frac{D_K(P_{\Theta^n|\pi}\| P_{\Theta^n})}{\log n} 
> \frac{\dim_{P_\Pi}(\pi)}{2}\right\},
\end{multline*}
we have proved so far that $\lim_{n\rightarrow\infty} P_\Pi(\Pi'_{n,\ep}) = 0$
for all $\ep > 0$. Now define $n_0(1) = 1$ and for all $m > 1$,
$$
n_0(m) = \min_{n_0}
\colon P_\Pi(\Pi'_{n,\frac1m}) \leq \frac1m\quad\forall n\geq n_0.
$$
Note that $\Pi_{n,\frac1{m+1}} \subseteq \Pi_{n,\frac1m}$ so $n_0(m)$ is an
increasing function of $m$. For all $n\geq 1$ define 
$m_0(n) = \max_n\colon n_0(m) \leq n$
(with $m_0(n) = \infty$ if there is no maximium). Note that 
$m_0(n) \rightarrow \infty$ and so $\frac1{m_0(n)}\in o(1)$. 
Let 
\begin{multline*}
\Pi'_n = \left\{\pi\colon\frac{D_K(P_{\Theta^n|\pi}\| P_{\Theta^n})}{\log n} 
< \frac{\dim_{P_\Pi}(\pi)}{2} - \frac1{m_0(n)} \right.\\
\left.\text{or} \quad
\frac{D_K(P_{\Theta^n|\pi}\| P_{\Theta^n})}{\log n} 
> \frac{\dim_{P_\Pi}(\pi)}{2}\right\}.
\end{multline*}
By definition $P_\Pi(\Pi_n) \leq \frac 1{m_0(n)}$, hence $P_\Pi(\Pi_n)
\rightarrow 0$. Thus 
$$
\frac{D_K(P_{\Theta^n|\pi}\| P_{\Theta^n})}{\log n}  \thickapprox 
\frac{\dim_{P_\Pi}(\pi)}{2} + o(1).
$$
$\Box$

%% file: bayesbias.bbl
\begin{thebibliography}{10}

\bibitem{sontagseig}
F.~Albertini and E.~Sontag.
\newblock For neural networks function, determines form.
\newblock {\em Neural Networks}, 1994.

\bibitem{BLW}
P.~Bartlett, P.~Long, and B.~Williamson.
\newblock {Fat-Shattering and the Learnability of Real-Valued Functions}.
\newblock In {\em Proccedings of the Seventh ACM Conference on Computational
  Learning Theory}, New York, 1994. ACM Press.

\bibitem{BiasLearn95}
J.~Baxter.
\newblock {A Model of Bias Learning}.
\newblock Technical Report LSE-MPS-97, London School of Economics, Centre for
  Discrete and Applicable Mathematics, November 1995.
\newblock Submitted to Journal of the ACM.

\bibitem{colt95}
J.~Baxter.
\newblock {L}earning {I}nternal {R}epresentations.
\newblock In {\em {P}roceedings of the {E}ighth {I}nternational {C}onference on
  {C}omputational {L}earning {T}heory}, Santa Cruz, California, 1995. ACM
  Press.

\bibitem{Berger85}
J.~O. Berger.
\newblock {\em {Statistical Decision Theory and Bayesian Analysis}}.
\newblock Springer-Verlag, New York, 1985.

\bibitem{Berger86}
J.~O. Berger.
\newblock {Multivariate Estimation: Bayes, Empirical Bayes, and Stein
  Approaches}.
\newblock {\em SIAM}, 1986.

\bibitem{Bridle}
J.~S. Bridle.
\newblock Probabilistic interpretation of feedforward classification network
  outputs, with relationships to statistical pattern recognition.
\newblock In F.~Fogelman-Soulie and J.~Herault, editors, {\em Neurocomputing:
  Algorithms, Architectures}. Springer Verlag, New York, 1989.

\bibitem{CB1}
B.~Clarke and A.~Barron.
\newblock {Information-Theoretic Asymptotics of Bayes Methods}.
\newblock {\em IEEE Transactions on Information Theory}, 36:453--471, 1990.

\bibitem{CT}
T.~M. Cover and J.~A. Thomas.
\newblock {\em Elements of Information Theory}.
\newblock John Wiley \& Sons, Inc., New York, 1991.

\bibitem{Good80}
I.~J. Good.
\newblock {Some History of the Hierarchical Bayesian Methodology}.
\newblock In J.~M. Bernado, M.~H.~D. Groot, D.~V. Lindley, and A.~F.~M. Smith,
  editors, {\em Bayesian Statistics II}. University Press, Valencia, 1980.

\bibitem{HausslerCOLT95}
D.~Haussler and M.~Opper.
\newblock {General Bounds on the Mutual Information Between a Parameter and n
  Conditionally Independent Observations}.
\newblock In {\em Proccedings of the Eighth ACM Conference on Computational
  Learning Theory}, New York, 1995. ACM Press.

\bibitem{kurkova}
V.~Kurkova and P.~C. Kainen.
\newblock Functionally equivalent feedforward neural networks.
\newblock {\em Neural Computation}, 6:543--558, 1994.

\bibitem{Mackay}
D.~Mackay.
\newblock {B}ayesian {I}nterpolation.
\newblock {\em Neural Computation}, 4:415--447, 1991.

\bibitem{Mackay2}
D.~Mackay.
\newblock {T}he {E}vidence {F}ramework {A}pplied to {C}lassification
  {N}etworks.
\newblock {\em Neural Computation}, 4:698--714, 1991.

\bibitem{sussman}
H.~J. Sussmann.
\newblock Uniqueness of the weights for minimal feedforward nets with a given
  input-output map.
\newblock {\em Neural Networks}, 5:589--594, 1992.

\bibitem{Valiant}
L.~G. Valiant.
\newblock A theory of the learnable.
\newblock {\em Comm. ACM}, 27:1134--1142, 1984.

\bibitem{VC2}
V.~N. Vapnik.
\newblock {\em Estimation of Dependences based on Empirical Data}.
\newblock Springer-Verlag, New York, 1982.

\bibitem{VC1}
V.~N. Vapnik and A.~Y. Chervonenkis.
\newblock On the uniform convergence of relative frequencies of events to their
  probabilities.
\newblock {\em Theory Probab. Appl.}, 16:264--280, 1971.

\end{thebibliography}
